%% file: main.tex
\documentclass[10pt,twocolumn,letterpaper]{article}

\usepackage{cvpr}              
\input{preamble}
\definecolor{cvprblue}{rgb}{0.21,0.49,0.74}
\usepackage[pagebackref,breaklinks,colorlinks,allcolors=cvprblue]{hyperref}

\usepackage{amssymb} 
\usepackage{multirow} 
\usepackage{booktabs}
\usepackage{fontawesome}
\usepackage{hyperref}
\usepackage[symbol]{footmisc}
\usepackage[accsupp]{axessibility}

\def\paperID{44551} 
\def\confName{CVPR}
\def\confYear{2026}

\title{PASR: Pose-Aware 3D Shape Retrieval from Occluded Single Views}
\newcommand{\OURS}{PASR\xspace}
\author{
\textbf{Jiaxin Shi}$^1$\textsuperscript{\faEnvelopeO}
\thanks{\textsuperscript{\faEnvelopeO} Corresponding author: Jiaxin Shi (\href{mailto:shijiaxinzzzqyh@gmail.com}{\text{shijiaxinzzzqyh@gmail.com}})},
\textbf{Guofeng Zhang}$^2$\thanks{Project Lead}, 
\textbf{Wufei Ma}$^2$,
\textbf{Naifu Liang}$^3$,
\textbf{Adam Kortylewski}$^4$,
\textbf{Alan Yuille}$^{2}$ \\
\small $^1$Shanghai Jiao Tong University \quad 
$^2$Johns Hopkins University \quad \\
\small $^3$ University of California, San Diego \quad 
$^4$ CISPA Helmholtz Center for Information Security \\
}

\begin{document}
\maketitle
\input{sec/0_abstract}    
\vspace{-4mm}
\input{sec/1_intro}
\input{sec/2_related_works}
\input{sec/3_method}
\input{sec/4_experiments}
\input{sec/5_conclusion}

\paragraph{Acknowledgments.}
This work was supported by the National Eye Institute (NEI) under Award R01EY037193 and the Office of Naval Research (ONR) under Grant N000142512132. The authors would also like to thank Jiawei Peng for helpful discussions and valuable assistance.

{
    \small
    \bibliographystyle{ieeenat_fullname}
    \bibliography{main}
}

\input{sec/X_suppl}


\end{document}

%% file: sec/0_abstract.tex
\begin{abstract}
Single-view 3D shape retrieval is a fundamental yet challenging task that is increasingly important with the growth of available 3D data. 
Existing approaches largely fall into two categories: those using contrastive learning to map point cloud features into existing vision-language spaces and those that learn a common embedding space for 2D images and 3D shapes. However, these feed-forward, holistic alignments are often difficult to interpret, which in turn limits their robustness and generalization to real-world applications. 
To address this problem, we propose Pose-Aware 3D Shape Retrieval (\OURS), a framework that formulates retrieval as a feature-level analysis-by-synthesis problem by distilling knowledge from a 2D foundation model (DINOv3) into a 3D encoder. By aligning pose-conditioned 3D projections with 2D feature maps, our method bridges the gap between real-world images and synthetic meshes.
During inference, \OURS performs a test-time optimization via analysis-by-synthesis, jointly searching for the shape and pose that best reconstruct the patch-level feature map of the input image. This synthesis-based optimization is inherently robust to partial occlusion and sensitive to fine-grained geometric details. \OURS substantially outperforms existing methods on both clean and occluded 3D shape retrieval datasets by a wide margin. Additionally, \OURS demonstrates strong multi-task capabilities, achieving robust shape retrieval, competitive pose estimation, and accurate category classification within a single framework.
\end{abstract}

%% file: sec/1_intro.tex
\section{Introduction}
\label{sec:intro}

\begin{figure}[htbp]
\centering
\includegraphics[width=1\linewidth]{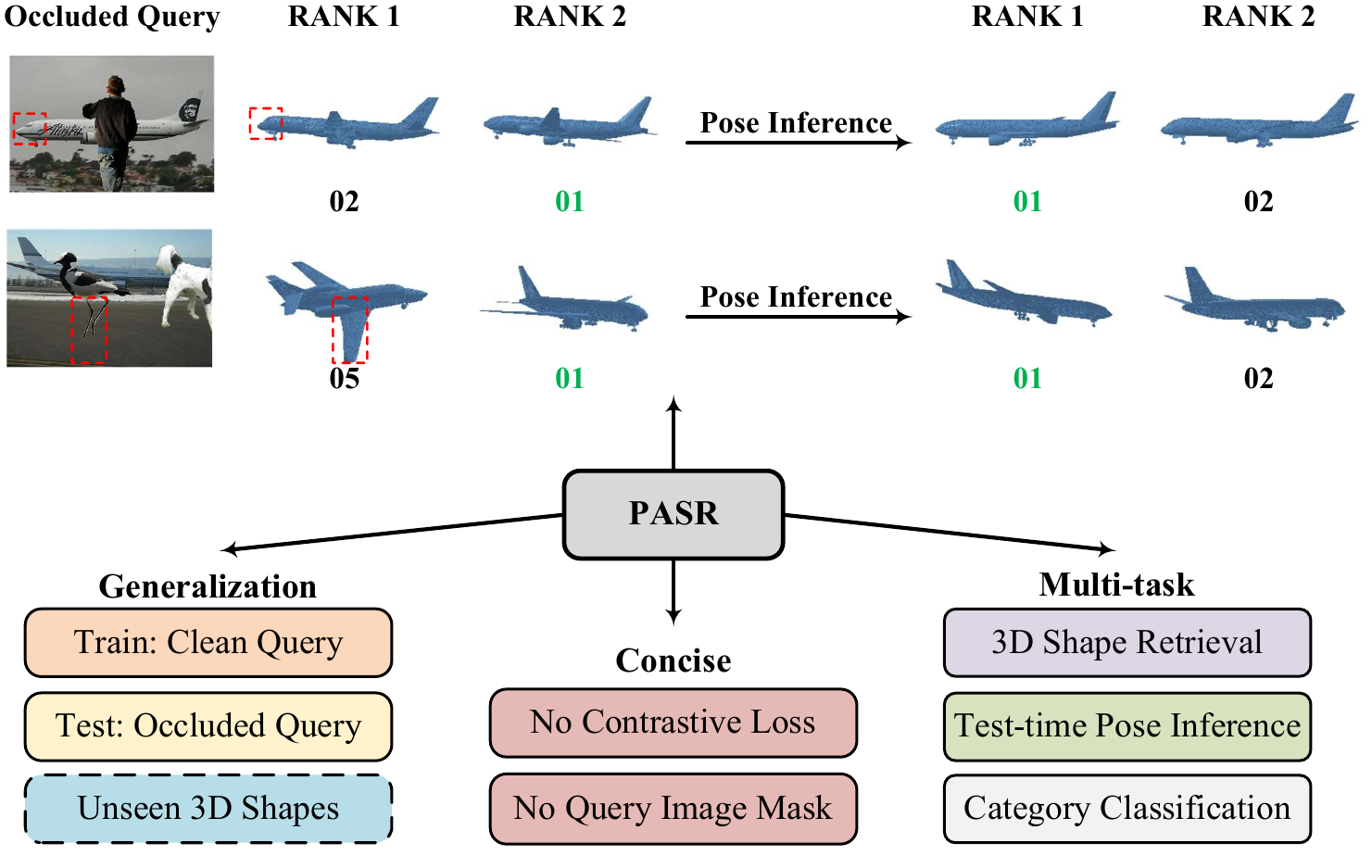}
\caption{
\OURS achieves competitive clean, occluded, and unseen 3D shape retrieval from single images, while also enabling multi-tasking with accurate category classification and pose estimation.
}
\label{fig:teaser}
\vspace{-4mm}
\end{figure}

A fundamental challenge in computer vision is enabling machines to perceive the 3D world from 2D images. Single-view 3D shape retrieval~\cite{tangelder2004survey,xiao2020survey} directly addresses this problem, requiring a model to retrieve a corresponding 3D mesh given only a single RGB image. As large-scale 3D data becomes increasingly common, the need for effective search methods becomes increasingly significant. More broadly, the ability to understand the 3D shape from a single view is a key enabler for inverse graphics~\cite{yuille2006vision}. This, in turn, allows for detailed 3D scene representations, which are critical for advancing downstream applications such as 3D world understanding~\cite{shi2025chain, ma2025spatialreasoner}, autonomous navigation~\cite{chen2025splat}, and robotic manipulation~\cite{yan2024robotron}.

Recent works have explored 3D shape retrieval on two main fronts. The first line of work has primarily advanced the field by leveraging large-scale multimodal alignment. It aligns 3D shape features with an existing image-text embedding space~\cite{liu2023openshape, zhou2023uni3d, xue2023ulip, xue2024ulip}. While these approaches excel on synthetic benchmarks, their generalization to real-world images is limited. In practical scenarios, available 3D models are often not an exact instance-level match for 2D images. Instead, a single generic 3D shape is frequently used as a proxy for many distinct 2D object instances. This creates a critical gap for this line of work since these methods require precise alignments of synthetic data and thus fail to generalize to real-world scenarios. The second line of work embeds 2D images and 3D shapes into a joint feature space~\cite{fu2020hard, wu2023generalizing, lin2021single, grabner2019location, songsc-ibsr, lin2021cmic}. They typically learn holistic, global embeddings rather than explicit 3D geometry representations. This view-agnostic design is inherently vulnerable to partial occlusions and limits generalization to unseen 3D mesh models. Furthermore, both lines of work fail to sufficiently explore rich 3D information, like pose, that is available during training, precluding their use for fine-grained, multi-task applications.

To address these limitations of prior work, we propose Pose-Aware 3D shape retrieval (\OURS), a novel framework that reformulates retrieval as a feature-level analysis-by-synthesis problem. As shown in Figure~\ref{fig:teaser}, our framework is designed for high robustness, particularly against partial occlusions, and exhibits strong generalization to unseen object meshes. \OURS operates in two stages. During training, we distill fine-grained knowledge from a 2D foundation model into our 3D encoder. This design of obtaining an explicit 3D representation is the key to its strong generalization capabilities to novel mesh shapes. During inference, \OURS performs a test-time optimization, jointly searching for the optimal shape and its 3D pose that best reconstructs the feature map of the query image. This synthesis-based alignment is inherently sensitive to fine-grained geometry and robust to occlusions, allowing our unified framework to excel beyond retrieval to some multi-task objectives, including 3D pose estimation and image classification. 

\OURS achieves state-of-the-art performance on both the Pix3D and Pascal3D datasets. In particular, \OURS attains an overall Top-1 retrieval accuracy of 81.59\% on Pix3D and 76.43\% on Pascal3D. Notably, \OURS exhibits strong robustness, especially when the query image is occluded. Compared with the previous best baselines, \OURS achieves average relative improvements of 11.09\% and 7.15\% on Pix3D and Pascal3D. Furthermore, our method shows better generalization to unseen 3D shapes.

Our contributions can be summarized as follows:
\begin{itemize}
\item We propose \OURS, a novel framework that learns explicit, point-level 3D representations of 3D objects from a 2D foundation model. This approach demonstrates high robustness against partial occlusion and achieves strong generalization to unseen meshes.

\item \OURS fully utilizes 3D information, enabling it to exhibit strong multi-tasking capabilities. It delivers competitive performance on 3D pose estimation and category classification, even in challenging occluded scenarios.

\item \OURS achieves state-of-the-art shape retrieval performance, outperforming previous methods on the Pix3D and Pascal3D benchmarks, with average relative improvements of 11.09\% and 7.15\%, respectively. 

\end{itemize}

%% file: sec/2_related_works.tex
\section{Related Works}
\label{sec:related_works}

\paragraph{3D shape retrieval.} 
Research in 3D shape retrieval mainly follows two directions: directly using 3D model representations for retrieval and employing image-based techniques to infer 3D shapes. 
Model-based methods~\cite{chen2003visual,han2019aggregating,jiang2019mlvcnn, 10818713} extract representative 3D features and measure feature similarities, achieving strong performance in retrieval and classification. However, they require a 3D query model, which is often impractical in real-world scenarios. 
Conversely, image-based retrieval methods represent 3D shapes as multi-view images~\cite{lin2021single,nguyen2022templates, hu2023cross, pal2024domain, song2023domain,song2025adaptive}, bridging the 2D–3D gap by embedding real RGB images and rendered views into a shared space via triplet or contrastive learning. 
Recent works~\cite{wu2023generalizing,chu2024open} aim to improve generalization and robustness under occlusion, yet feed-forward methods still struggle in such cases. 
To address this, render-and-compare strategies have been explored. Our proposed \OURS follows this paradigm, iteratively refining predictions by comparing with rendered images, yielding more robust retrieval under challenging occluded conditions.

\vspace*{-0.2cm}
\paragraph{Feature-level analysis-by-synthesis.} Previous feature-level analysis-by-synthesis methods~\cite{kortylewski2020compositional} learn neural representations of objects, which enables localizing occlusions for robust object detection. Later works~\cite{wang2021nemo,ma2022robust, jesslen2024novum, Yuan_2025_ICCV} extended this approach to 3D neural mesh models by learning neural features for each vertex of the 3D shapes and performing pose estimation with render-and-compare. However, these 3D compositional models are often limited to a single template shape per category due to the part-level contrastive loss adopted in training. This drawback in model design also limits the ability of these models to represent more fine-grained shapes. Our method builds on but significantly extends standard feature-level analysis-by-synthesis methods by learning point-level feature representations and replacing the part-contrastive loss with feature-level reconstruction loss. Our \OURS features a simpler architecture design and allows us to jointly represent the 3D pose, 3D shape, and visibility of an object.

\vspace*{-0.2cm}
\paragraph{2D-3D knowledge distillation.}
Some works have explored building 3D foundation models to reproduce the success of 2D foundation models~\cite{xue2023ulip, xue2024ulip, liu2023openshape, zhou2023uni3d, hegde2023clip, Peng2023OpenScene, wu2024ppt,wu2025sonata}. Due to the limited availability of 3D data and computational resources, some methods attempt to leverage 2D foundation models to distill knowledge into 3D space, facilitating more generalized shape representations. These methods can be generally categorized into two types. The first uses contrastive learning to align the 3D embeddings of 3D models with the 2D embeddings of 2D foundation models~\cite{xue2023ulip,xue2024ulip,liu2023openshape,zhou2023uni3d,hegde2023clip}. Although these methods succeed in some 3D tasks, they remain constrained by the limited quality and diversity of available 3D training data, unlike the data available in the 2D domain. The second method bypasses the training of 3D-specific models by directly applying 2D encoders to rendered views of 3D objects or scenes~\cite{partslip,zhang2021pointclip,Zhu2022PointCLIPV2,zhang2025concerto}. Some studies render 3D shapes from multiple viewpoints, process these views with 2D foundation models, and subsequently use the extracted features for downstream tasks. Recent studies have also attempted to back-project and aggregate 2D features onto 3D shapes~\cite{wimmer2024back}. Although these methods share similarities with \OURS, the key difference lies in our direct rendering of point-level 3D features from the 3D model into 2D space, followed by distillation of 2D knowledge within 2D space. Our method preserves 2D spatial features more effectively than previous methods. In addition, \OURS retains pose-conditioned information, which is advantageous for retrieval and other viewpoint-dependent downstream tasks.

%% file: sec/3_method.tex
\section{Method}
\label{sec:method}
\begin{figure*}[htbp]
\centering
\includegraphics[width=1\linewidth]{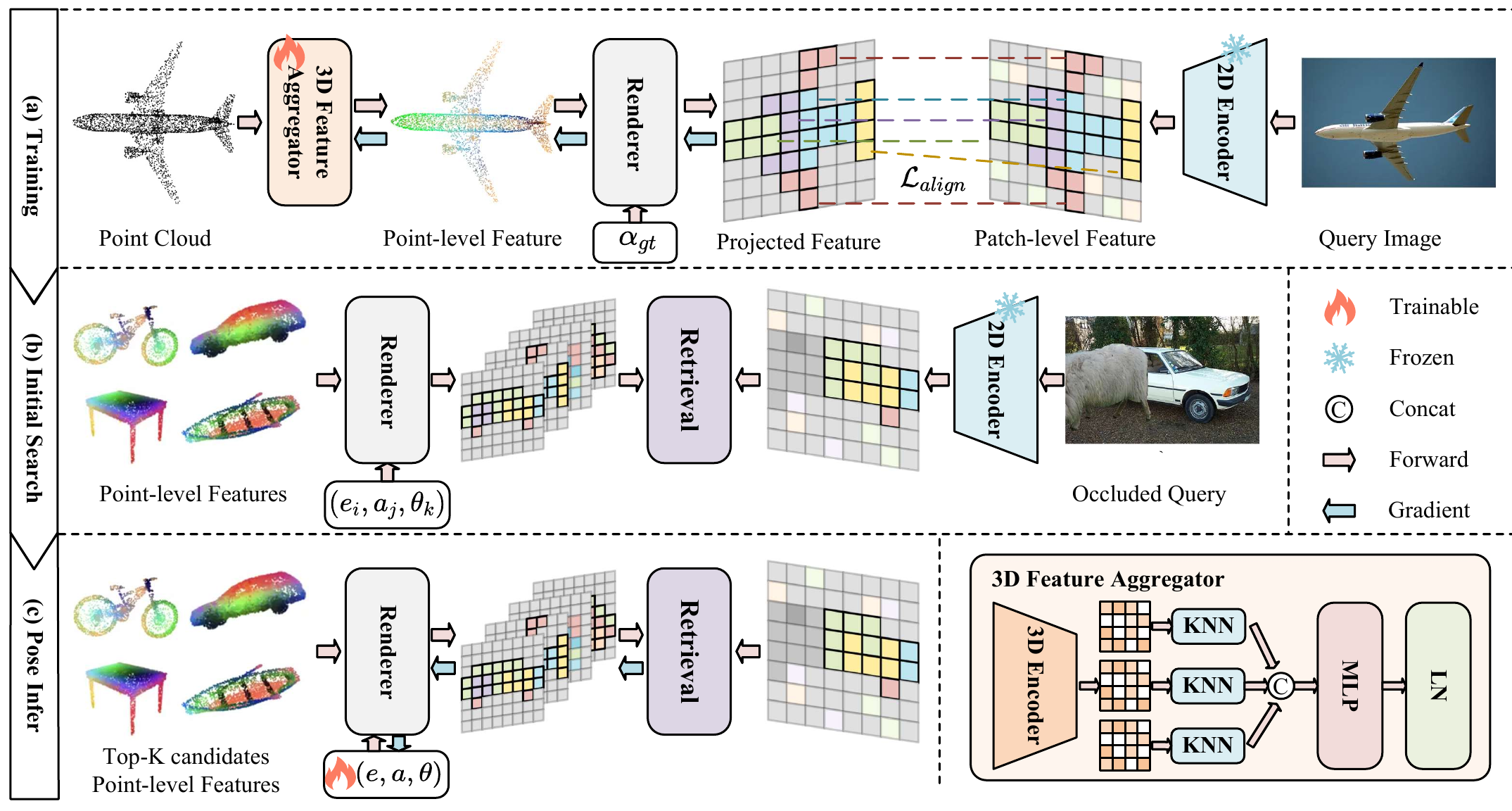}
\caption{Overall architecture of \OURS. (a) We distill semantic and spatial knowledge from a 2D foundation model into a 3D point cloud encoder. (b) Given a query image, we first perform an initial search by rendering the features of each 3D shape from a set of predefined poses and comparing them to the query image features. (c) Then, for the top-k candidates, we iteratively refine their poses to find the shape and pose that best reconstruct the query's feature map via analysis-by-synthesis.}
\label{fig:overall}
\vspace*{-0.2cm}
\end{figure*}

We introduce \OURS, a pose-aware 3D shape retrieval framework that formulates single-view 3D shape retrieval as a feature-level analysis-by-synthesis problem. Below, we first outline the problem formulation in Sec. \ref{subsec:preliminaries}. Then we detail how we learn
rich, view-aware 3D point features by distilling knowledge in Sec. \ref{subsec:alignment}. Finally, we describe how \OURS performs a test-time optimization via analysis-by-synthesis to achieve multi-tasking in Sec. \ref{subsec:retrieval}. The overall architecture of \OURS is illustrated in Figure \ref{fig:overall}.

\subsection{Preliminaries}
\label{subsec:preliminaries}
In single-view 3D shape retrieval, we are given a real-world query image $I$. The image can exhibit complex colors, textures, backgrounds, lighting conditions, and varying degrees of occlusion. Given a large database of 3D shapes $\mathcal{D}=\left\{\mathcal{T}_1, \mathcal{T}_2, \dots\right\}$, our goal is to retrieve the shape $\mathcal{T}^* \in \mathcal{D}$ that best matches the object in $I$. We first sample each shape $\mathcal{T}_i$ into a point cloud $\mathcal{P}$ of $C$ points, denoted $\mathcal{P}=\left\{p_j \right\}_{j=1}^{C} \in \mathbb{R}^{C \times 3}$. 
Since the viewpoint in $I$ is unknown, retrieval starts by comparing the query image feature map with 2D feature maps $f_p$ rendered from each candidate 3D shape under various poses. These rendered maps are projections of learned point-level features, which we denote as $\mathcal{M}$. To enable this comparison, we employ a differentiable renderer $\mathcal{R}$ that projects the 3D features $\mathcal{M}$ into the 2D feature space according to a given camera pose $\alpha$.
We use a pretrained 2D foundation model to extract a feature map $f_{q} \in \mathbb{R}^{D \times H \times W}$. Here, $D$ is the dimension of the feature, and $H$ and $W$ are the height and width of the feature map. During training, we leverage the ground truth pose $\alpha_{gt}$ to supervise the alignment between $f_{q}$ and the corresponding projected map of point-level features. 
At inference, for each candidate shape $\mathcal{T}_i$, we optimize its pose $\alpha$ to find the best alignment, thereby jointly determining the best matching 3D shape and its corresponding pose.

\subsection{Training Setup}
\label{subsec:alignment}
\paragraph{3D feature extraction.}
We first extract a set of descriptive, point-level features $\mathcal{M} = \{m_1, \dots, m_C\}$ for a given point cloud $\mathcal{P} = \{p_1, \dots, p_C\}$. This set contains a $D$-dimensional feature vector $m_i$ for each corresponding point $p_i \in \mathcal{P}$. We adopt a 3D encoder $\Phi$ to extract multi-scale features. At a given scale $s$, the encoder outputs a downsampled point set $\mathcal{P}_s$ and its corresponding features $\mathcal{M}_{s}$:
\begin{equation}
\left(\mathcal{P}_s, \mathcal{M}_{s}\right)=\Phi_s\left(\mathcal{P}\right),
\end{equation}
where $\mathcal{M}_{s}$ is the set of $D_s$-dimensional feature vectors for the $N_s$ points in $\mathcal{P}_s$. To capture both local geometric detail from fine scales and global semantic context from coarse scales, we select outputs from a set of scales $s \in \mathcal{S}$. We then interpolate these features from each $\mathcal{M}_{s}$ back to the original point cloud resolution $\mathcal{P}$. We first find the $k$ nearest neighbors in $\mathcal{P}_s$ for each $p_i \in \mathcal{P}$:
\begin{equation}
\mathcal{N}_i^{(s)}= \text{KNN}(p_i, \mathcal{P}_s, k).
\end{equation}

We then apply a temperature-scaled softmax weighting over the neighbor distances as 

\begin{equation}
d_{i,k}^{(s)}=\left\| p_i-p_k^{(s)}\right\|_2^2,~~ p_k^{(s)} \in \mathcal{N}_i^{(s)}
\end{equation}

\begin{equation}
w_{i,k}^{(s)}= \frac{\exp \left(-d_{i,k}^{(s)}/ \tau\right)}{\sum_{k' \in \mathcal{N}_i^{(s)}}\exp\left(-d_{i,k'}^{(s)}/\tau \right)},
\end{equation}
where $\tau > 0$ controls the sharpness. The interpolated feature $\hat{m}_{s,i}$ for $p_i$ at scale $s$ is the weighted average of neighbor features $m_k^{(s)} \in \mathcal{M}_s $ as
\begin{equation}
\hat{m}_{s,i} = \sum_{k \in \mathcal{N}_i^{(s)}}w_{i,k}^{(s)} \cdot m_{k}^{(s)}, \quad 
\end{equation}

This process yields a set of interpolated features $\hat{\mathcal{M}}_s = \{\hat{m}_{s,1}, \dots, \hat{m}_{s,C}\}$ for each scale $s$. Finally, we obtain the full set of multi-scale, point-level features by concatenating the interpolated features from all selected scales $\mathcal{S}$ for each point. These are passed through an MLP layer with layer normalization as
\begin{equation}
\mathcal{M} = MLP\left(\text{Concat}_{s \in \mathcal{S}}\left(\hat{\mathcal{M}}_{s} \right) \right),
\end{equation}
where $\mathcal{M}$ denotes the final point-level 3D features, containing one $D$-dimensional vector $m_j$ for each point $p_j \in \mathcal{P}$. This multi-scale aggregation ensures that each feature $m_j$ extracted by the 3D encoder contains both fine-grained local geometric structure and global semantic context.

\vspace*{-0.2cm}
\paragraph{2D-3D feature alignment.}
Our objective is to distill knowledge from the 2D foundation model into the 3D encoder. 
To achieve this, we enforce alignment in the 2D feature space. Specifically, we project the learned 3D point features $\mathcal{M}$ onto a feature map.
We employ a differentiable point-based renderer $\mathcal{R}$ with ground-truth camera pose $\alpha_{gt}$ to generate a projected 2D feature map $f_p$ as
\begin{equation}
f_p=\mathcal{R}\left(\mathcal{P}, \mathcal{M}, \alpha_{gt}\right),
\end{equation}
where $f_p \in \mathbb{R}^{D \times H \times W}$ has the same dimensions as the 2D foundation model's output, enabling direct comparison.

We train $\Phi$, our 3D encoder by aligning the projected features $f_p$ with the query features $f_q$ using a cosine similarity loss. Our renderer provides a foreground mask $M \in \{0,1\}^{H \times W}$ to ensure alignment only on the object's pixels. Notably, we do not require a ground-truth segmentation mask of the query image during training and inference. The loss $\mathcal{L}_{align}$ is the masked average of the cosine similarity,
\begin{equation}
    \mathcal{L}_{align}=\frac{\sum_{u,v}\left(1-\mathcal{S}\left(f_{q}^{(u,v)}, ~f_p^{(u,v)}\right) M^{(u,v)}\right)}{\sum_{u,v} M^{(u,v)}}.
\end{equation}
where $\mathcal{S}$ denotes cosine similarity. This alignment strategy effectively distills the rich semantic knowledge of the 2D foundation model into our 3D encoder.

\subsection{3D shape retrieval}
\label{subsec:retrieval}
\paragraph{Initial search.}
During inference, the ground-truth pose $\alpha_{gt}$ is unknown. We therefore perform an initial search to find a good starting pose for the optimization and to coarsely rank the candidate 3D shapes from the database $\mathcal{D}$. We compare the 2D query image feature $f_q$ against each projected feature $f_p$, rendered under a discretized set of $N$ predefined candidate poses. This pose set is generated by uniformly sampling the elevation, azimuth, and in-plane rotation angles as  $\mathcal{E}=\left\{e_1, \dots,e_{N_e}\right\}$, $\mathcal{A}=\left\{a_1, \dots, a_{N_a}\right\}$, $\Theta=\left\{\theta_1, \dots, \theta_{N_\theta}\right\}$, where $N_e$, $N_a$, $N_{\theta}$ are the number of samples for each angle.

For a given candidate shape $\mathcal{T}$, we first extract its 3D features $\mathcal{M}$ using our trained encoder $\Phi$. Then, for each pose candidate $\alpha_n$ in our discrete set, we project its features to 2D using our renderer: $f_p^{(n)} = \mathcal{R}(\mathcal{P}, \mathcal{M}, \alpha_n)$. We measure the similarity between the query $f_q$ and each projection $f_p^{(n)}$ using our alignment loss $\mathcal{L}_{align}$ as a score. The pose $\alpha_{init}$ from our discrete set that yields the minimum loss for a given shape is selected as its best coarse alignment, and the value of this loss serves as its initial ranking score.

\vspace*{-0.2cm}
\paragraph{Test-time pose inference for top-k candidates.}

The accuracy of our retrieval heavily depends on finding the correct camera pose that aligns the candidate shape with the query image. Therefore, we introduce a test-time refinement step for the top-k candidates identified in the initial search.
We iteratively update the poses of the top-k 3D shape candidates, minimizing $\mathcal{L}_{align}$ using the AdamW optimizer.
To ensure efficient optimization and simplicity in bounding the search space, we optimize the intuitive Euler angles $(e, a, \theta)$ instead of directly optimizing the rotation matrix $R \in SO(3)$, which is a constrained manifold.
After several iterations of test-time pose refinement, we obtain the re-ranked top-k candidates and their optimal poses. Then we calculate the final alignment score for $\mathcal{T}_i$ and the query image $I$ with
\begin{equation}
\mathcal{L}\left(\mathcal{T}_i,I\right)=\mathcal{L}_{align}\left(f_{q},f_{p}^i\right), 
\end{equation}
where $f_{p}^i$ is the projected 2D feature map generated by the candidate $\mathcal{T}_i \in \left\{\mathcal{T}_1, \dots, \mathcal{T}_k\right\}$ and its optimal 3D pose. The 3D shape with minimum alignment loss is the result:
\begin{equation}
    \mathcal{T}^*=\arg \min_{\mathcal{T}_i}\mathcal{L}(\mathcal{T}_i,I).
\end{equation}

This two-stage retrieval method significantly improves efficiency. By pre-extracting point-level features and limiting the test-time pose inference to only the top-k candidates, \OURS  strikes an effective balance between accuracy and computational cost. 

In addition, we optimize only three parameters for each candidate, allowing efficient batch processing. 

%% file: sec/4_experiments.tex
\section{Experiment}
\label{sec:exp}

\subsection{Setup}

\paragraph{Datasets.} We evaluate our method on the Pix3D~\cite{sun2018pix3d} and Pascal3D~\cite{xiang2014beyond} datasets. Pix3D contains four categories (bed, chair, sofa, and table). For Pascal3D, we use all twelve categories for training and evaluation. We follow the preprocessing pipeline in prior works such as Nemo~\cite{wang2021nemo} and ImageNet3D~\cite{ma2024imagenet3d} to generate occluded images. We define three levels of occlusion, occluding 2-10\% (L1), 10-20\% (L2), and 20-40\% (L3) of the object's area. To ensure fairer comparisons and better evaluate the model's robustness under varying occlusion conditions, query image masks were not used in any of the experiments.

\vspace*{-0.2cm}
\paragraph{Baselines.}
We mainly focus on the 3D shape retrieval task. 
We compare our results to several baseline models for single-image 3D shape retrieval and 3D pose estimation. For 3D shape retrieval, we select CMIC~\cite{lin2021single}, SC-IBSR~\cite{songsc-ibsr}, and recent unified 3D shape models, such as OpenShape~\cite{liu2023openshape} and Uni3D~\cite{zhou2023uni3d}. For pose estimation, we compare our method against ResNet-50~\cite{he2016deep} and Swin-T~\cite{liu2021swin}. We retrained CMIC and SC-IBSR using their official codebases and original data augmentation settings. For the large-scale pre-trained models, such as OpenShape and Uni3D, we directly evaluated their publicly released checkpoints. For pose estimation, ResNet-50 and Swin-T accept only the query images as inputs and are optimized using the ground truth 3D poses and predict the pose in a single forward pass at test time. In contrast, our method performs test-time pose inference by optimizing the pose of each candidate shape to minimize a feature alignment loss between the query image and the projected features of 3D candidates. 

\vspace*{-0.2cm}
\paragraph{Implementation details.} Our model is trained for 300 epochs with a batch size of 8 using 4 NVIDIA RTX Titan GPUs. We utilize AdamW to optimize the model; the learning rate is 1e-4, and the weight decay is 1e-2. Considering inference speed and alignment performance, we sample 4096 points from a mesh and uniformly sample 192 predefined poses (4 elevations, 12 azimuths, and 4 in-plane rotations) for the initial search stage. We adopt DINOv3~\cite{simeoni2025dinov3} as the 2D encoder and PointNeXt~\cite{qian2022pointnext} as the 3D encoder. During the point-level feature extraction stage, we use multi-scale features from PointNeXt. The set of selected scales, denoted by $\mathcal{S}$, corresponds to the final three layers, which have 512, 256, and 128 feature dimensions, respectively. We set the number of nearest neighbors $K=3$, the temperature $\tau=5e-2$. We use PyTorch3D to render the 3D features into 2D space. The camera intrinsics and extrinsics follow the settings provided by the datasets. For test-time pose inference, we also use AdamW as the optimizer; the learning rate is 0.01, the batch size is 64, and the weight decay is 1e-2. We only optimize three parameters, $elev$, $azim$, and $\theta$, optimizing them for 50 steps. 

\subsection{Comparison to Prior Works}
\paragraph{Single image 3D shape retrieval.}
Table \ref{tab:pix3d_retrieval} and Table \ref{tab:pascal3d_retrieval} illustrate the single-image 3D shape retrieval results on Pix3D and Pascal3D under different levels of occlusion, respectively. On Pix3D, our method outperforms the other baselines overall and in every category. Zero-shot methods, OpenShape and Uni3D, do not perform as well as the supervised models. While the performance of all methods degrades sharply under heavy (L3) occlusion, our method still outperforms the strongest baseline, CMIC, by 5.94\%. On Pascal3D, our method also outperforms the other compared models. Especially in occlusion scenarios, our method significantly outperforms the comparison methods. For L3 occlusion, our approach achieves 8.29\% higher accuracy than CMIC and 17.21\% higher than Uni3D.

Results on Pix3D and Pascal3D demonstrate that our method can achieve highly competitive scores without employing contrastive losses to train the model to distinguish between different instances of the same category. This is accomplished only by aligning 2D and 3D features in 2D space and performing test-time pose inference. 

\begin{table}[t]
\centering
\small
\setlength{\tabcolsep}{4pt}
\begin{tabular}{l|c|cccc|c}
\toprule
\textbf{Model} & \textbf{Level} & \textbf{Bed} & \textbf{Chair} & \textbf{Sofa} & \textbf{Table} & \textbf{Overall} \\
\midrule
\multirow{4}{*}{OpenShape} 
& L0 & 36.17 & 33.45 & 29.44 & 67.05 & 37.57 \\
& L1 & 33.51 & 21.87 & 18.89 & 55.68 & 26.92 \\
& L2 & 31.38 & 16.19 & 18.33 & 50.85 & 22.75 \\
& L3 & 28.72 & 10.86 & 13.89 & 38.35 & 16.80 \\
\midrule
\multirow{4}{*}{Uni3D} 
& L0 & 36.70 & 34.75 & 44.44 & 59.09 & 40.49 \\
& L1 & 25.53 & 23.96 & 25.37 & 41.48 & 26.88 \\
& L2 & 24.47 & 19.86 & 19.44 & 40.06 & 23.00 \\
& L3 & 21.28 & 15.11 & 20.37 & 36.36 & 19.76 \\
\midrule
\multirow{4}{*}{CMIC} 
& L0 & 63.30 & 85.90 & 67.60 & 66.50 & 77.40 \\
& L1 & 52.70 & 79.40 & 61.10 & 57.70 & 70.30 \\
& L2 & 47.30 & 71.40 & 55.00 & 50.90 & 63.00 \\
& L3 & 34.60 & 53.20 & 36.30 & 36.90 & 45.80 \\
\midrule
\multirow{4}{*}{SC-IBSR} 
& L0 & 62.77 & 82.30 & 64.81 & 64.49 & 74.45 \\
& L1 & 59.04 & 75.97 & 57.41 & 55.11 & 67.65 \\
& L2 & 48.40 & 69.06 & 48.33 & 45.45 & 59.60 \\
& L3 & 35.64 & 51.29 & 32.96 & 30.97 & 43.20 \\
\midrule
\multirow{4}{*}{Ours} 
& L0 & \textbf{63.83} & \textbf{88.29} & \textbf{74.81} & \textbf{75.00} & \textbf{81.59} \\
& L1 & \textbf{61.70} & \textbf{83.96} & \textbf{72.41} & \textbf{74.43} & \textbf{78.38} \\
& L2 & \textbf{57.45} & \textbf{78.16} & \textbf{66.85} & \textbf{64.20} & \textbf{72.13} \\
& L3 & \textbf{52.66} & \textbf{55.24} & \textbf{47.04} & \textbf{44.60} & \textbf{51.74} \\
\bottomrule
\end{tabular}
\vspace*{-0.2cm}
\caption{3D shape retrieval results on Pix3D (L0-L3).}
\label{tab:pix3d_retrieval}
\end{table}

\vspace*{-0.2cm}
\paragraph{3D pose estimation.}
Since 3D pose estimation is central to our shape retrieval framework, we also evaluate the pose estimation accuracy $Acc_{\pi/18}$ of our test-time pose inference process. Table \ref{tab:pose_estimation} presents pose metrics for our method and comparison baselines. Unlike prior competing methods, our method first retrieves the top-k 3D shapes using a set of predefined poses, and then re-ranks these results by optimizing the poses for each candidate. During 3D pose inference, our method does not know the ground truth top-1 3D shape. Therefore, we only use the pose of the top-1 retrieved 3D shape for metric calculation. On Pix3D and Pascal3D, although our method is not designed for pose estimation and only performs inference at test time, our method outperforms the learning-based baselines under L0, L1, L2, and L3 occlusion. For example, on Pix3D, our approach achieves 70.0\% at L0 and 58.0\% at L1, surpassing the second-best ResNet-50 by a large margin. Similarly, on the Pascal3D dataset, our method obtains the best performance, reaching 40.3\% at L0 and 36.7\% at L1. These results highlight the generality and robustness of our method for pose estimation.

\begin{table}[t]
\centering
\small
\begin{tabular}{l|cccc|c}
\toprule
\textbf{Level} & \textbf{Uni3D} & \textbf{OpenShape} & \textbf{CMIC} & \textbf{SC-IBSR} & \textbf{Ours} \\
\midrule
L0 & 54.01 & 55.80 & 75.44 & 66.42 & \textbf{76.43} \\
L1 & 50.25 & 49.80 & 70.68 & 58.65 & \textbf{73.21} \\
L2 & 46.73 & 47.99 & 65.87 & 50.58 & \textbf{71.49} \\
L3 & 45.84 & 42.56 & 54.76 & 38.12 & \textbf{63.05} \\
\bottomrule
\end{tabular}
\vspace*{-0.3cm}
\caption{3D shape retrieval results on Pascal3D (L0-L3).}
\label{tab:pascal3d_retrieval}
\vspace*{-0.1cm}
\end{table}

\begin{table}[t]
\centering
\small
\setlength{\tabcolsep}{4pt}
\begin{tabular}{@{}l *{8}{c}@{}}
\toprule
\multirow{2}{*}{\textbf{Method}} & \multicolumn{4}{c}{$\textbf{Acc}_{\pi/18} \uparrow$ (\textbf{Pix3D})} & \multicolumn{4}{c}{$\textbf{Acc}_{\pi/18} \uparrow$ (\textbf{Pascal3D})} \\
\cmidrule(lr){2-5} \cmidrule(l){6-9}
 & L0 & L1 & L2 & L3 & L0 & L1 & L2 & L3 \\
\midrule
Swin-T      & 39.4 & 30.6 & 25.0 & 19.3 & 29.6 & 26.2 & 22.8 & 17.7 \\
ResNet-50   & 48.1 & 40.4 & 30.9 & 17.7 & 37.6 & 32.6 & 28.6 & 20.7 \\
\textbf{Ours} & \textbf{70.0} & \textbf{58.0} & \textbf{41.3} & \textbf{19.9} & \textbf{40.3} & \textbf{36.7} & \textbf{34.3} & \textbf{25.7} \\
\bottomrule
\end{tabular}
\vspace*{-0.3cm}
\caption{Performance comparison on Pix3D and Pascal3D under rotation granularity $\pi/18$.}
\label{tab:pose_estimation}
\vspace*{-0.1cm}
\end{table}

\begin{table}[t]
\centering
\small
\setlength{\tabcolsep}{2pt}
\begin{tabular}{@{}l *{8}{c}@{}}
\toprule
\multirow{2}{*}{\textbf{Method}} & \multicolumn{4}{c}{\textbf{Pix3D}} & \multicolumn{4}{c}{\textbf{Pascal3D}} \\
\cmidrule(lr){2-5} \cmidrule(l){6-9}
 & L0 & L1 & L2 & L3 & L0 & L1 & L2 & L3 \\
\midrule
OpenShape & 93.85 & 92.06 & 90.77 & 86.15 & 97.32 & 97.62 & 96.55 & 94.15 \\
Uni3D     & 94.82 & 90.69 & 89.88 & 86.48 & 95.83 & 94.89 & 93.81 & 92.45 \\
CMIC      & 97.33 & 95.91 & 93.16 & 86.19 & 98.08 & 96.17 & 93.10 & 83.40 \\
SC-IBSR   & 96.64 & 95.14 & 91.26 & 82.71 & 94.40 & 88.93 & 82.38 & 66.23 \\
\textbf{Ours} & \textbf{98.95} & \textbf{98.01} & \textbf{97.61} & \textbf{93.81} & \textbf{99.28} & \textbf{99.11} & \textbf{98.72} & \textbf{96.70} \\
\bottomrule
\end{tabular}
\vspace*{-0.3cm}
\caption{Category-level classification performance comparison (\%) of different methods on Pix3D and Pascal3D datasets across levels L0–L3. Higher accuracy is better.}
\label{tab:cls_performance}
\vspace*{-0.2cm}
\end{table}

\begin{figure*}[t]
\centering
\includegraphics[width=1\linewidth]{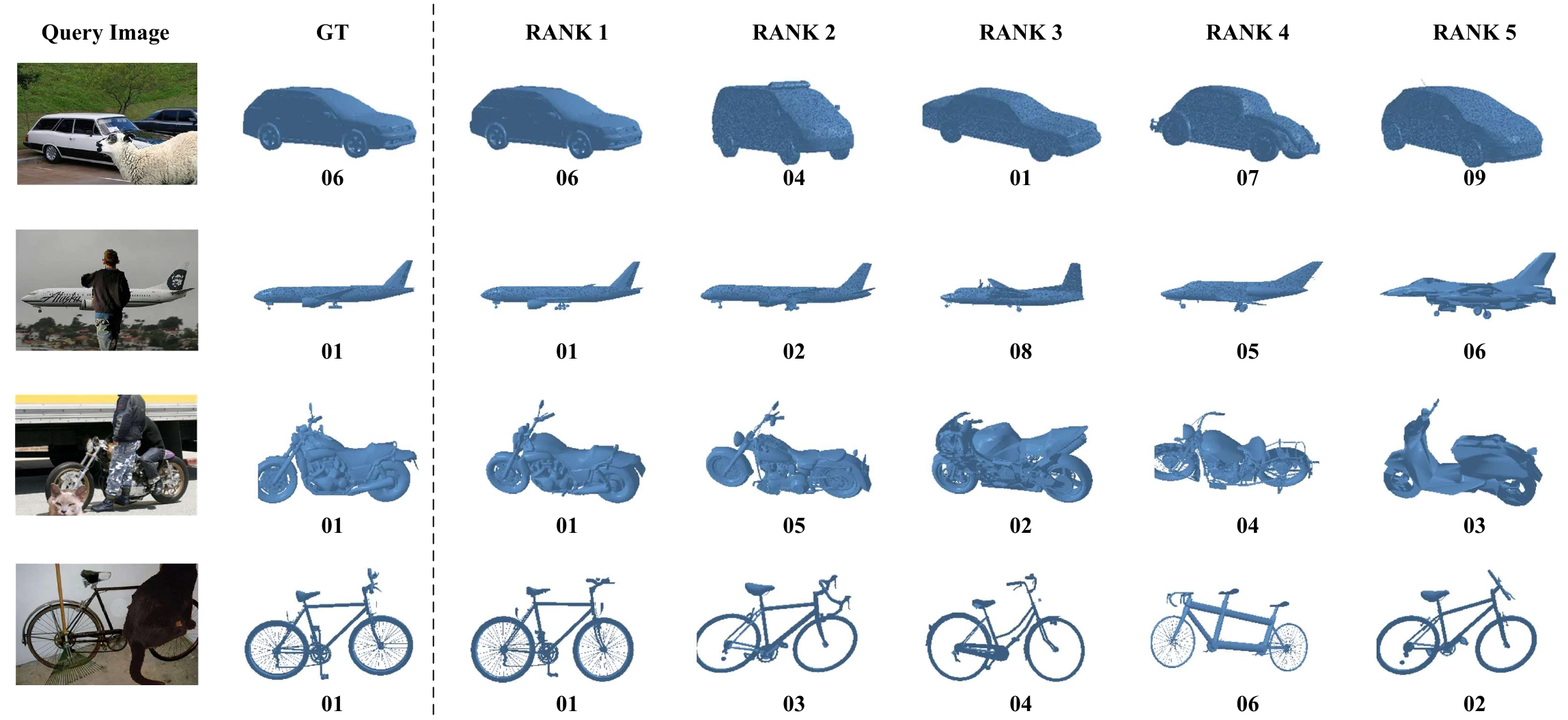}
\caption{Retrieval results on Pascal3D (L3). The "GT" column presents the ground truth 3D shape and its corresponding ground truth 3D pose. While retrieving 3D shapes, our method can estimate pose for every candidate.}
\label{fig:retrieval}
\vspace*{-0.2cm}
\end{figure*}

\begin{figure}[htbp]
\centering
\includegraphics[width=1\linewidth]{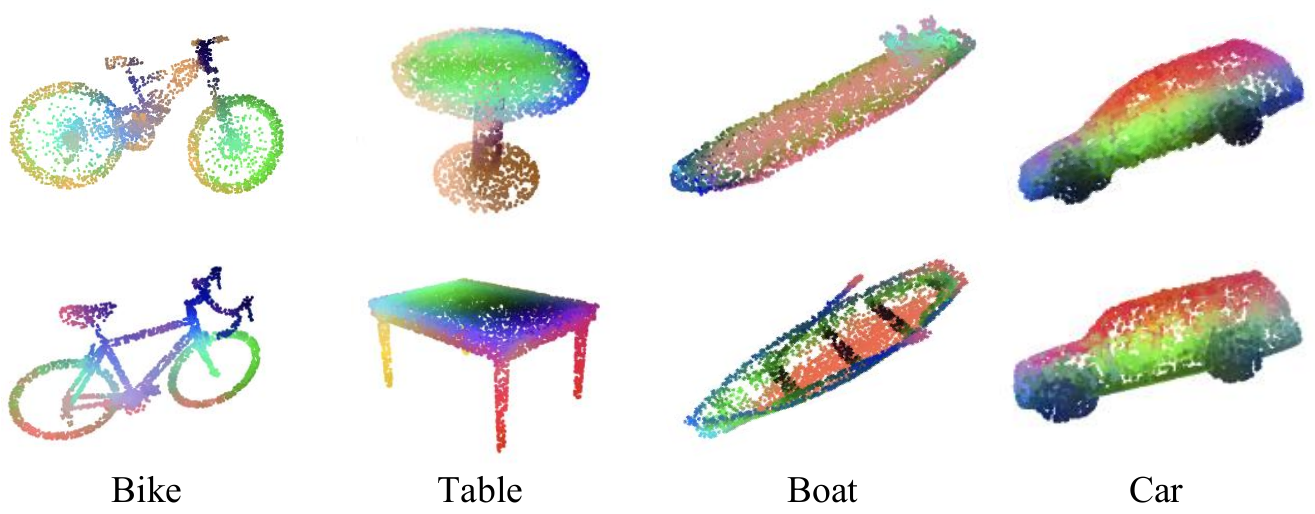}
\caption{Visualization of learned point-level features from 3D encoder via PCA.}
\label{fig:pca_pointcloud}
\vspace*{-0.3cm}
\end{figure}

\vspace*{-0.2cm}
\paragraph{Category-level classification.}
Category-level classification is the foundation for instance-level retrieval. Thus, we evaluate this metric on Pix3D and Pascal3D. As shown in Table~\ref{tab:cls_performance}, our method outperforms all compared baselines on Pix3D and Pascal3D with different occlusion levels. Prior unified 3D shape models, i.e., OpenShape and Uni3D, demonstrated strong zero-shot category-level classification performance, achieving results comparable to models trained on these two datasets. Specifically, on Pascal3D L3 occlusion, OpenShape and Uni3D outperformed CMIC by 10.75\% and 9.05\%, respectively. 

However, despite these unified 3D models' strong category-level discrimination, they struggle to distinguish different instances within the same category. In contrast, our approach excels at differentiating distinct instances while maintaining robustness in category-level classification tasks. The performance drop of our approach from L0 to L3 on Pix3D and Pascal3D is merely 5.14\% and 2.58\%, respectively. In comparison, CMIC exhibits L3 classification accuracy on Pix3D and Pascal3D that is 11.14\% and 14.68\% lower than L0. 

\vspace*{-0.5cm}
\paragraph{Transfer to unseen 3D shapes.}
To further evaluate robustness, we test the ability of our method and the supervised baselines to transfer to unseen 3D shapes. We evaluate these models that are trained on Pix3D, and we test their performance on Pascal3D (L0) without any fine-tuning. We choose the chair category for evaluation, which is common to both datasets. Notably, the 3D shapes from the Pix3D and Pascal3D datasets are disjoint.

As shown in Table~\ref{tab:zero_shot}, our approach significantly outperforms baselines. CMIC achieves 85.90\% top-1 retrieval accuracy (L0) on the chair category of Pix3D. However, when transferred to Pascal3D, it only achieves 9.13\% Top-1 retrieval accuracy. Similar results were also obtained for SC-IBSR. In comparison, our method can still achieve better performance. Our method achieves 63.40\% Top-1 retrieval accuracy and 91.70\% Top-5 retrieval accuracy. Using test-time pose inference achieves a 5.82\% higher Top-1 retrieval accuracy compared to using only the initial search. These results demonstrate that projecting 3D features into 2D space for comparison with the query image features, followed by test-time pose inference, achieves better generalization performance for unseen 3D shapes.

\begin{table}[t]
\centering
\small
\begin{tabular}{l|ccc}
\toprule
\textbf{Method} & \textbf{Top-1} $\uparrow$ & \textbf{Top-5} $\uparrow$ & \textbf{CLS} $\uparrow$\\
\midrule
CMIC & 9.13 & 31.94 & 52.09 \\
SC-IBSR & 4.18 & 38.02 & 71.10 \\
Ours$^\dagger$ & 57.58 & 91.70 & - \\
Ours & \textbf{63.40} & \textbf{91.70} & \textbf{93.18} \\
\bottomrule
\end{tabular}
\vspace*{-0.3cm}
\caption{Comparison of transfer ability on unseen 3D shapes (chair) on Pascal3D (L0). CLS stands for category-level Top-1 classification. $^\dagger$ indicates that \OURS uses only the initial search without test-time pose inference. Higher accuracy (\%) is better.}
\label{tab:zero_shot}
\vspace*{-0.6cm}
\end{table}

\subsection{Qualitative Results}

\paragraph{Retrieval \& pose estimation.}
Figure \ref{fig:retrieval} illustrates qualitative retrieval results of our method on Pascal3D (L3). Our method not only distinguishes well among similar candidates but also accurately estimates their poses. Notably, even the other mismatched candidates can be optimized to a suitable pose, making them more aligned with the query image. Additionally, the object in the query image suffers from occlusion, which increases the difficulty of extracting useful features and reduces the clues available for inference. Our method enables dynamic pose inference during test time, helping the model better identify the correct 3D pose and thereby distinguish differences between various candidate 3D shapes more effectively.

\begin{table}[t]
\centering
\small
\setlength{\tabcolsep}{3pt}
\begin{tabular}{ccccccccc}
\toprule
 & \multicolumn{4}{c}{\textbf{Pix3D}} & \multicolumn{4}{c}{\textbf{Pascal3D}} \\
\cmidrule(lr){2-5} \cmidrule(lr){6-9}
 & L0 & L1 & L2 & L3 & L0 & L1 & L2 & L3 \\
\midrule
Init. & 76.78 & 73.40 & 65.86 & 48.87 & 66.97 & 64.95 & 61.46 & 55.68 \\
Opt.  & \textbf{81.59} & \textbf{78.38} & \textbf{72.13} & \textbf{51.74} & \textbf{76.43} & \textbf{73.21} & \textbf{71.49} & \textbf{63.05} \\
\midrule
$\Delta$ & +4.81 & +4.98 & +6.27 & +2.87 & +9.46 & +8.26 & +10.03 & +7.37 \\
\bottomrule
\end{tabular}
\vspace*{-0.2cm}
\caption{Ablation study on using test-time pose inference. Top-1 retrieval accuracy comparison between Pix3D and Pascal3D across occlusion levels L0-L3.}
\label{tab:ablation_pose_inference}
\vspace*{-0.2cm}
\end{table}

\begin{table}[t]
\centering
\small
\begin{tabular}{ccccc}
\toprule
\textbf{Step} & \textbf{Top-1 $\uparrow$} & \textbf{ACC$_{\pi/6} \uparrow$} & \textbf{ACC$_{\pi/18} \uparrow$} & \textbf{MedErr $\downarrow$} \\
\midrule
0 & 55.68 & - &  - & - \\
\midrule
10 & 59.58 & 52.32 & 21.29 & 27.34 \\
20 & 61.90 & 54.86 & 26.02 & 24.54 \\
30 & 62.20 & 54.81 & \textbf{26.89} & 24.57 \\
\textbf{50} & \textbf{63.05} &\textbf{55.18} & 25.72 & \textbf{23.85} \\
\bottomrule
\end{tabular}
\vspace*{-0.2cm}
\caption{Ablation study on the number of steps for test-time pose inference for Pascal3D (L3). }
\label{tab:pascal3d_l3_step_ablation}
\vspace*{-0.5cm}
\end{table}

\vspace*{-0.4cm}
\paragraph{Point-level feature visualization.}
Figure \ref{fig:pca_pointcloud} demonstrates that our feature alignment strategy successfully lifts 2D knowledge to the 3D encoder. The learned point-level features represent fine-grained semantic and location information. For instance, they distinguish between the wheels and the frame of the bike, the left and right sides of the table, and the front and rear of the car and boat. The consistent color patterns of the features across objects within the same category highlight the robustness of our learned 3D point-level representation. These features are crucial for the analysis-by-synthesis stage, as they provide precise local cues for aligning the 3D shape with the object in the query image.

\subsection{Ablation Study}

\paragraph{Test-time pose inference.} 
We conduct two ablation studies on test-time pose inference. We first compare the top-1 retrieval performance between using test-time pose inference and using initial search only. Second, we ablate the number of pose optimization steps.

As shown in Table~\ref{tab:ablation_pose_inference}, using test-time pose inference can significantly improve the effectiveness of our method. Furthermore, Table \ref{tab:pascal3d_l3_step_ablation} details the impact of the number of pose inference steps. In order to demonstrate the performance of our method for occlusion cases, we show the results on Pascal3D (L3). At the test time, we only set three parameters, i.e., $elev$, $azim$, and $\theta$ for optimization. Step 0 means not using test-time pose inference; thus, this row shows the initial search result. With only 10 steps of pose inference, the performance improves by 3.9\% compared to the initial search. More optimization steps are beneficial for retrieval and pose estimation performance but also increase inference time. Therefore, we choose 50 steps as a balance between performance and efficiency.

\vspace*{-0.3cm}
\paragraph{Number of predefined poses.}
As shown in Table \ref{tab:view_config_comparison}, setting more predefined poses can improve initial search performance. Increasing the number of views from 24 to 648 increases top-1 accuracy by 12.06\%. More views result in longer inference times during the initial search phase. Considering the balance between effectiveness and efficiency, our main experiment employs only 192 views.

\begin{table}[t]
\centering
\small

\begin{tabular}{cccc}
\toprule
$elev\times azim \times \theta$ & \textbf{Views} & \textbf{Top-1 $\uparrow$} & \textbf{Top-5 $\uparrow$} \\
\midrule
2$\times$6$\times$2 & 24  & 56.13 & 95.83 \\
3$\times$9$\times$3 & 81  & 63.84 & 98.64 \\
4$\times$12$\times$4 & 192 & 66.40 & 98.62 \\
6$\times$18$\times$6 & 648 & \textbf{68.19} & \textbf{98.89} \\
\bottomrule
\end{tabular}
\vspace*{-0.2cm}
\caption{Retrieval performance comparison across different numbers of predefined poses on Pascal3D (L0) in initial search stage.}
\label{tab:view_config_comparison}
\vspace*{-0.5cm}
\end{table}

\begin{table}[t]
\centering
\small
\setlength{\tabcolsep}{3pt}
\begin{tabular}{c|c|cccc}
\toprule
Occ. & 2D Encoder & \textbf{Top-1 $\uparrow$} & \textbf{ACC$_{\pi/6} \uparrow$} & \textbf{ACC$_{\pi/18} \uparrow$} & \textbf{MedErr $\downarrow$}\\
\midrule
\multirow{3}{*}{L0}
 & DINOv3 & \textbf{76.43} & 80.20 & 40.29 & 12.07  \\
 & DINOv2 & 70.14 & 80.79 & \textbf{44.21} & \textbf{11.37}  \\
 & DIY-SC & 68.30 & \textbf{82.06} & 42.59 & 11.74  \\
\midrule
\multirow{3}{*}{L3}
 & DINOv3 & 63.05 & \textbf{55.18} & 25.72 & 23.85  \\
 & DINOv2 & \textbf{64.05} & 68.83 & \textbf{34.16} & \textbf{15.29}  \\
 & DIY-SC & 57.99 & 68.53 & 28.40 & 16.61  \\
\bottomrule
\end{tabular}
\vspace*{-0.2cm}
\caption{Ablation study on 2D encoder on Pascal3D (L0, L3).}
\label{tab:ablation_2d_encoder}
\vspace*{-0.2cm}
\end{table}

\begin{table}[t]
\centering
\small
\begin{tabular}{l|cccc}
\toprule
& \textbf{Top-1 $\uparrow$} & \textbf{ACC$_{\pi/6} \uparrow$} & \textbf{ACC$_{\pi/18} \uparrow$} & \textbf{MedErr $\downarrow$}\\
\midrule
Multi-scale & \textbf{63.05} & \textbf{55.18} & \textbf{25.72} & \textbf{23.85}  \\
Last Layer  & 61.29 & 53.71 & 20.94 & 25.70  \\
2nd Last    & 58.84 & 51.28 & 19.53 & 28.38 \\
\bottomrule
\end{tabular}
\vspace*{-0.2cm}
\caption{Ablation study on the $i$-th layer from the end of the 3D encoder on Pascal3D (L3).}
\label{tab:ablation_3d_encoder}
\vspace*{-0.5cm}
\end{table}

\vspace*{-0.2cm}
\paragraph{2D encoder and 3D encoder.} We ablate the 2D encoder on Pascal3D (L0, L3) as shown in Table~\ref{tab:ablation_2d_encoder}. Generally, DINOv3 performs well at various occlusions. Even when replacing DINOv3 to other 2D encoders, \textit{i.e.,} DINOv2~\cite{oquab2023dinov2} and DIY-SC~\cite{duenkel2025diysc}, \OURS still outperforms the previous best baseline CMIC, which demonstrates our analysis-by-synthesis framework is very robust in high occlusion scenarios (L3). Table~\ref{tab:ablation_3d_encoder} shows ablation on $i$-th layer from the 3D encoder on Pascal3D (L3). Aggregating features from different layers is beneficial for obtaining multi-scale 3D features.

\subsection{Efficiency}
\OURS is a coarse-to-fine framework. Test-time pose inference is optional for enhancing the robustness at occlusion scenarios, and it only optimizes the top-k candidates, no matter how many shapes they have at the initial search stage. At test time, the 3D point-level features are pre-computed offline. For a query image, the heavy 2D encoder is executed only once. \OURS only renders the point features and optimizes the pose of each candidate.

%% file: sec/5_conclusion.tex
\section{Conclusion}
\label{sec:conclusion}

We have presented \OURS, a novel pose-aware framework for single-view 3D shape retrieval that reformulates retrieval as a feature-level analysis-by-synthesis problem. Unlike prior feed-forward methods that rely on static and view-agnostic embeddings, \OURS explicitly models pose-conditioned point-level 3D representations, thus enabling robust reasoning over occluded data at test time. By distilling 2D knowledge from a 2D foundation model into a 3D encoder, our method bridges the 2D–3D domain gap and learns geometry-aware, fine-grained 3D features. Through comprehensive experiments on Pix3D and Pascal3D datasets, \OURS achieves state-of-the-art performance, improving robustness under varying occluded query images and unseen 3D shapes. Its unified design naturally extends to multiple tasks in a single framework, such as category-level classification and pose estimation, without additional supervision. 
{\parfillskip=20pt
Future work will focus on generalizing this framework to diverse meshes across more categories and exploiting its capabilities for more downstream 3D tasks in the real world. 
\par
}

%% file: sec/X_suppl.tex
\clearpage
\setcounter{page}{1}
\maketitlesupplementary

\setcounter{section}{0}   
\renewcommand{\thesection}{\Alph{section}}

\section{Additional Implementation Details}

\paragraph{2D Encoder.}
We adopt the \texttt{vit\_base\_patch16} variant of DINOv3~\cite{simeoni2025dinov3} as our 2D encoder. We use the output of the final transformer layer as the 2D feature representation. The input query images are resized to a resolution of \(3 \times 592 \times 592\), and the resulting feature maps have a spatial resolution of \(37 \times 37\) with 768 channels, \textit{i.e.}, the encoder output has shape \(768 \times 37 \times 37\).

\paragraph{Renderer.}
We employ the point-based renderer from PyTorch3D, using \texttt{PointRasterizer} followed by \texttt{NormWeightedCompositor} to project point-level features into the 2D image plane. The rendered feature maps have a spatial resolution of \(37 \times 37\), matching the resolution of the 2D encoder output. The point radius is set to 0.04, and the number of points per pixel is set to 16. The focal length and principal point are adjusted according to the output resolution, and normalized device coordinates (NDC) are disabled. The background color is set to black.

\paragraph{Data Preprocessing.}

\begin{figure}[t]
\centering
\includegraphics[width=1\linewidth]{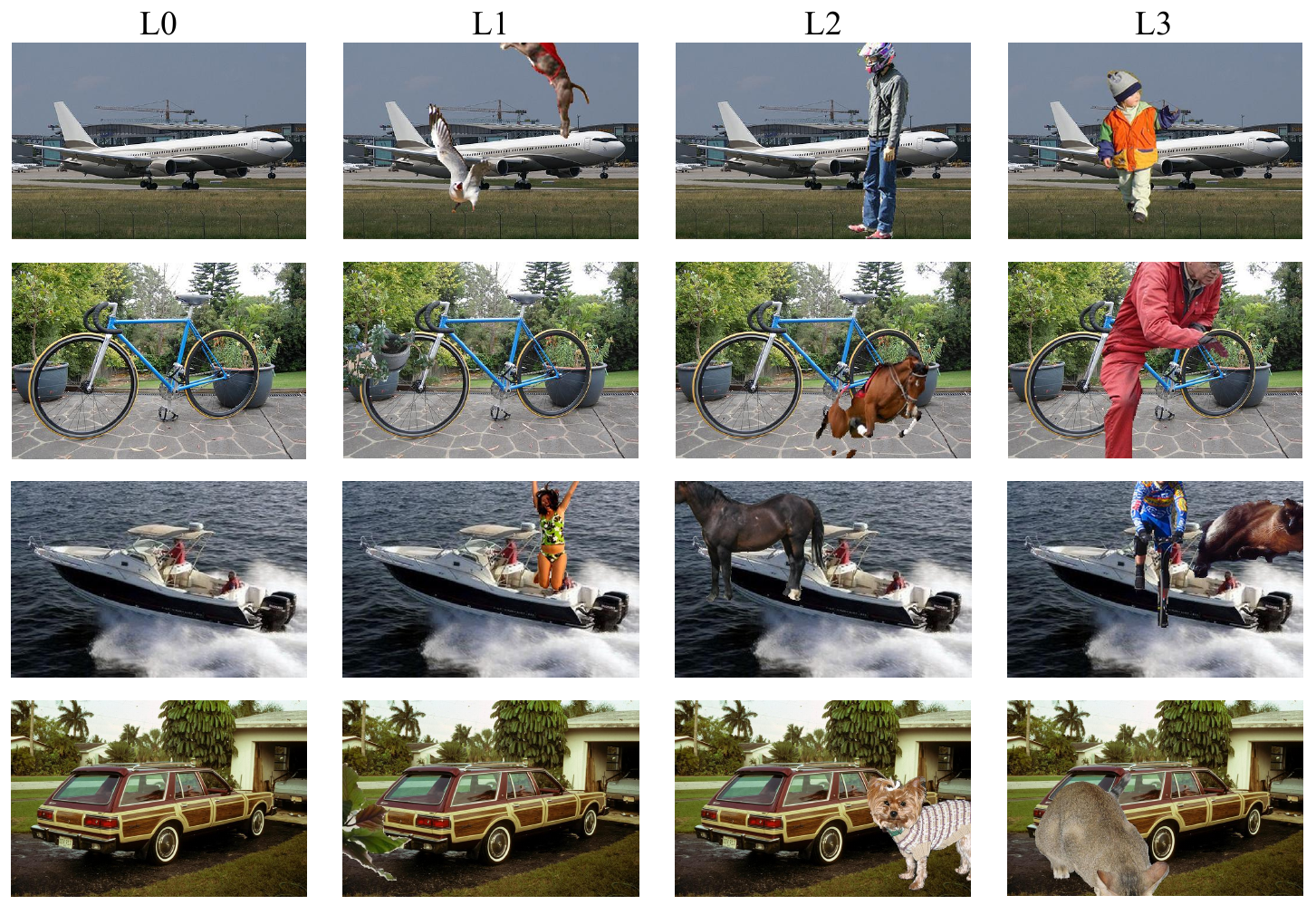}
\caption{Examples of the query images across different occlusion levels (L0-L3).}
\label{fig:query_images}
\vspace{-2mm}
\end{figure}

For 3D shapes, we first convert the meshes into point clouds using PyTorch3D and then normalize the point clouds.
Specifically, we uniformly sample 4096 points on the mesh surface, with sampling probability proportional to the face area, to obtain a batch of point clouds from a batch of meshes. For the CMIC~\cite{lin2021single} and SC-IBSR~\cite{songsc-ibsr} baselines, we apply standard data augmentation, including horizontal flipping, random cropping, and color transfer. For PASR, OpenShape~\cite{liu2023openshape}, and Uni3D~\cite{zhou2023uni3d}, we do not use any data augmentation. Figure~\ref{fig:query_images} shows the original occluded images. All input images are centered and resized to a resolution suitable for downstream 3D pose estimation, ensuring consistent scale and alignment across the dataset.

\section{Additional Results}
In this section, we provide additional results on Pix3D~\cite{sun2018pix3d} and Pascal3D~\cite{xiang2014beyond}.
Table \ref{tab:supp_pose_estimation} presents all pose metrics for our method and comparison baselines, Swin-T and ResNet-50. On Pix3D, our method outperforms the baselines under L0, L1, and L2 occlusion. Under severe occlusion (L3), the pose estimation accuracy is inherently bounded by the retrieval quality. Since our method retrieves the 3D shape first, incorrect shape retrieval in extreme occlusion scenarios leads to higher pose errors, resulting in performance lower than the baselines in this specific setting. On Pascal3D, our method achieves the best performance or highly competitive scores across all metrics and occlusion levels compared to Swin-T and ResNet-50. 

As shown in Table~\ref{tab:supp_pascal3d_retrieval}, on Pascal3D, our method outperforms the other compared models overall and in most categories. Notably, for some categories with a large number of samples, our method is minimally affected by occlusion. For instance, in the aeroplane category, our method only saw a decrease of 8.52\% from L0 to L3, whereas other competitive models, OpenShape and CMIC decreased by 39.5\% and 22.5\%, respectively. In the car category, our method only saw a 5.14\% drop from L0 to L3 as well, whereas CMIC decreased by 26.6\%, and SC-IBSR decreased by 34.14\%. We observe a counter-intuitive performance gain, such as Uni3D in the aeroplane category from L0 to L1. Upon qualitative inspection, we hypothesize that this is because the L0 images in Pascal3D often contain significant background clutter. The synthetic occlusion introduced in L1 may inadvertently suppress these distracting background features, forcing the attention of the model towards the discriminative geometric parts of the aeroplane (e.g., wings/fuselage), to which the ViT backbone is robust.

\begin{table*}[t]
\centering
\small
\setlength{\tabcolsep}{4pt}
\begin{tabular}{@{}l *{12}{c}@{}}
\toprule
\multirow{2}{*}{Method} & \multicolumn{4}{c}{$Acc_{\pi/6} \uparrow$} & \multicolumn{4}{c}{$Acc_{\pi/18} \uparrow$} & \multicolumn{4}{c}{MedErr $\downarrow$} \\
\cmidrule(lr){2-5} \cmidrule(lr){6-9} \cmidrule(l){10-13}
 & L0 & L1 & L2 & L3 & L0 & L1 & L2 & L3 & L0 & L1 & L2 & L3 \\
\midrule
\multicolumn{13}{@{}l}{$\nabla$\textbf{Pix3D}} \\
\midrule
Swin-T      & 72.0 & 66.0 & 60.8 & \textbf{50.6} & 39.4 & 30.6 & 25.0 & 19.3 & 12.5 & 16.1 & 19.4 & \textbf{29.0} \\
ResNet-50   & 80.4 & 73.4 & 61.8 & 43.9 & 48.1 & 40.4 & 30.9 & 17.7 & 10.3 & 12.5 & 17.3 & 41.4 \\
\textbf{Ours} & \textbf{89.1} & \textbf{82.8} & \textbf{66.7} & 38.6 & \textbf{70.0} & \textbf{58.0} & \textbf{41.3} & \textbf{19.9} & \textbf{6.5} & \textbf{8.4} & \textbf{13.5} & 50.3 \\
\midrule
\multicolumn{13}{@{}l}{$\nabla$\textbf{Pascal3D}} \\
\midrule
Swin-T      & 63.6 & 59.8 & 56.4 & 51.3 & 29.6 & 26.2 & 22.8 & 17.7 & 17.6 & 20.7 & 23.6 & 28.8 \\
ResNet-50   & 71.8 & 67.2 & 63.5 & \textbf{56.0} & 37.6 & 32.6 & 28.6 & 20.7 & 13.7 & 15.6 & 18.6 & 24.6 \\
\textbf{Ours} & \textbf{80.2} & \textbf{76.2} & \textbf{68.2} & 55.2 & \textbf{40.3} & \textbf{36.7} & \textbf{34.3} & \textbf{25.7} & \textbf{12.1} & \textbf{13.3} & \textbf{15.4} & \textbf{23.9} \\
\bottomrule
\end{tabular}
\vspace{-0.2cm}
\caption{Performance comparison on Pix3D and Pascal3D under different rotation granularities ($\pi/6$, $\pi/18$) and median error (MedErr).}
\label{tab:supp_pose_estimation}
\vspace{-0.3cm}
\end{table*}

\begin{table*}[t]
\centering
\small
\setlength{\tabcolsep}{4pt}
\begin{tabular}{l|c|cccccccccccc|c}
\toprule
Method & Level & Aero & Bike & Boat & Bottle & Bus & Car & Chair & Table & MBike & Train & Sofa & TV & Overall \\
\midrule
\multirow{4}{*}{Uni3D}
& L0 & 64.66 & 27.81 & 94.36 & 55.43 & 59.93 & 58.08 & 61.98 & 27.21 & 48.20 & 20.43 & 20.75 & 71.71 & 54.01 \\
& L1 & 77.75 & 24.06 & 91.09 & 51.90 & 52.81 & 47.65 & 59.32 & 24.96 & 51.48 & 16.67 & 13.52 & 65.79 & 50.25 \\
& L2 & 49.48 & 24.37 & 88.36 & 49.46 & 52.06 & 46.48 & 51.71 & 22.70 & 51.48 & 20.43 & 22.96 & 61.18 & 46.73 \\
& L3 & 63.20 & 19.69 & 84.55 & 44.84 & 52.81 & 45.08 & 45.25 & 22.88 & 50.82 & 11.29 & 27.04 & 54.28 & 45.84 \\
\midrule
\multirow{4}{*}{OpenShape}
& L0 & 82.12 & 32.19 & 82.91 & 39.67 & 37.08 & 45.67 & 53.61 & 46.97 & 71.15 & 75.27 & 65.09 & 52.96 & 55.80 \\
& L1 & 49.90 & 33.12 & 75.64 & 55.71 & 31.46 & 42.07 & 48.67 & 37.44 & 53.77 & 91.40 & 60.69 & 47.70 & 49.80 \\
& L2 & 46.36 & 28.12 & 74.91 & 49.73 & 38.95 & 39.94 & 44.49 & 28.08 & 65.25 & 92.47 & \textbf{70.13} & 37.83 & 47.99 \\
& L3 & 42.62 & 24.69 & 65.45 & 46.20 & 17.60 & 37.81 & 42.59 & 26.52 & 57.38 & 86.56 & \textbf{59.43} & 29.61 & 42.56 \\
\midrule
\multirow{4}{*}{CMIC}
& L0 & 78.80 & 56.90 & 89.10 & 68.20 & 81.30 & \textbf{72.70} & 67.70 & \textbf{73.10} & 82.30 & 96.20 & \textbf{71.10} & 77.00 & 75.44 \\
& L1 & 74.20 & 51.20 & 84.50 & 65.50 & 77.50 & 65.60 & 62.40 & \textbf{69.50} & 77.00 & 93.00 & \textbf{67.90} & 75.70 & 70.68 \\
& L2 & 66.50 & 52.50 & 81.10 & 59.00 & 71.50 & 58.40 & 52.90 & \textbf{67.60} & 75.70 & 91.40 & 64.50 & 72.00 & 65.87 \\
& L3 & 56.30 & \textbf{42.20} & 67.30 & 47.80 & 61.80 & 46.10 & 45.60 & \textbf{62.20} & 61.00 & 83.30 & 56.60 & 52.00 & 54.76 \\
\midrule
\multirow{4}{*}{SC-IBSR}
& L0 & 67.57 & 42.50 & 83.27 & 54.08 & 72.28 & 62.85 & 58.94 & 63.95 & 73.11 & 92.47 & 66.67 & 73.36 & 66.42 \\
& L1 & 54.05 & 35.94 & 72.00 & 49.73 & 68.16 & 53.96 & 48.29 & 60.31 & 66.56 & 87.63 & 62.26 & 65.46 & 58.65 \\
& L2 & 47.19 & 29.69 & 65.64 & 45.92 & 61.42 & 41.78 & 32.70 & 53.73 & 61.64 & 80.11 & 55.66 & 61.18 & 50.58 \\
& L3 & 33.26 & 18.44 & 46.36 & 39.13 & 44.19 & 28.71 & 25.86 & 47.66 & 55.74 & 73.12 & 38.68 & 40.13 & 38.12 \\
\midrule
\multirow{4}{*}{Ours}
& L0 & \textbf{89.19} & \textbf{57.50} & \textbf{96.74} & \textbf{74.73} & \textbf{86.14} & 70.41 & \textbf{75.67} & 59.00 & \textbf{86.56} & \textbf{97.85} & 62.58 & \textbf{84.87} & \textbf{76.43} \\
& L1 & \textbf{89.19} & \textbf{54.69} & \textbf{96.01} & \textbf{73.37} & \textbf{84.64} & \textbf{67.40} & \textbf{69.96} & 47.92 & \textbf{84.59} & \textbf{97.85} & 57.86 & \textbf{82.24} & \textbf{73.21} \\
& L2 & \textbf{87.87} & \textbf{52.81} & \textbf{95.39} & \textbf{70.16} & \textbf{82.86} & \textbf{66.28} & \textbf{67.89} & 45.63 & \textbf{83.61} & \textbf{96.77} & 56.74 & \textbf{79.47} & \textbf{71.49} \\
& L3 & \textbf{80.67} & 36.25 & \textbf{92.57} & \textbf{63.04} & \textbf{71.54} & \textbf{65.27} & \textbf{52.09} & 30.62 & \textbf{72.46} & \textbf{96.24} & 37.42 & \textbf{60.53} & \textbf{63.05} \\
\bottomrule
\end{tabular}
\vspace{-0.2cm}
\caption{Per-category 3D shape retrieval accuracy (\%) across occlusion levels L0–L3 on Pascal3D.}
\label{tab:supp_pascal3d_retrieval}
\vspace{-0.3cm}
\end{table*}

\section{Additional Qualitative Results}
\begin{figure*}[htbp]
\centering
\includegraphics[width=0.8\linewidth]{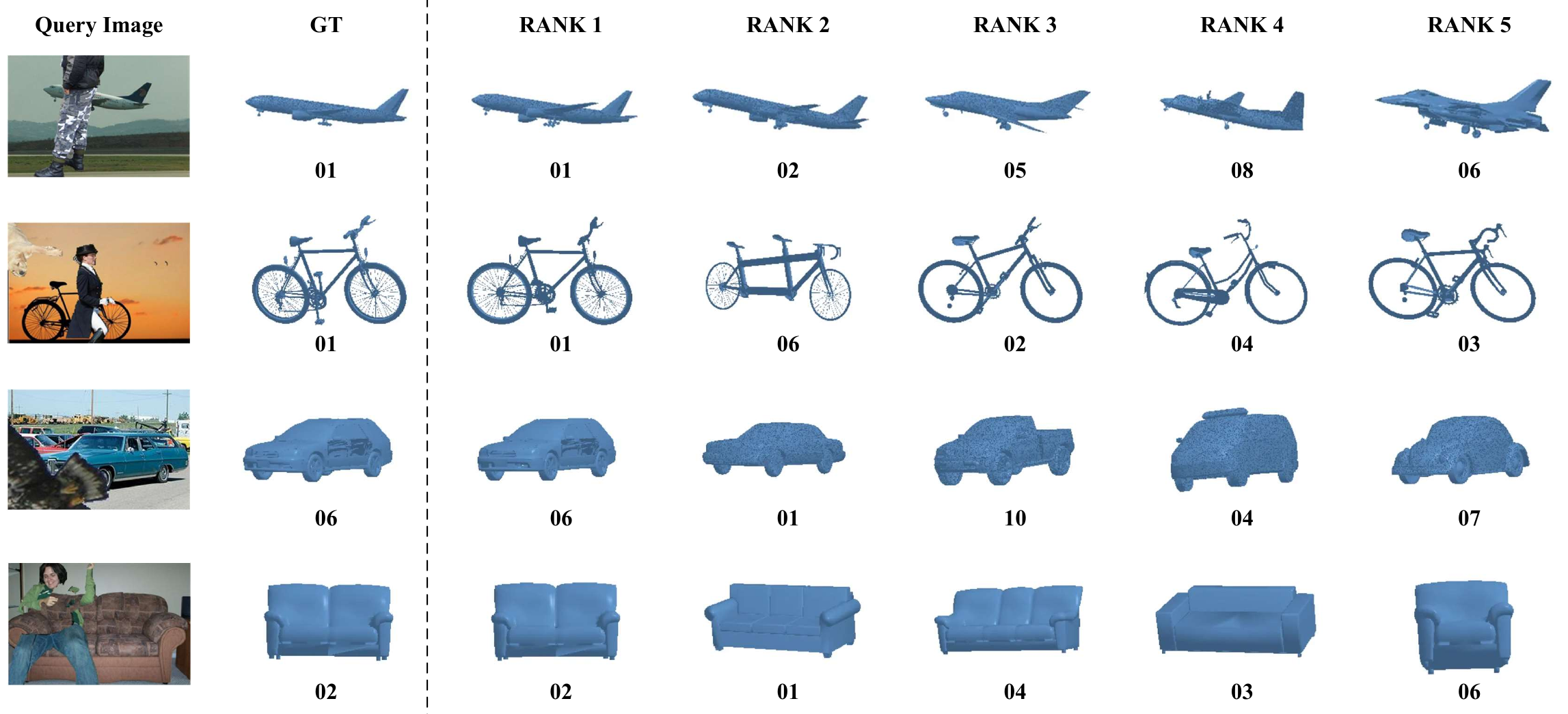}
\vspace{-0.2cm}
\caption{Additional 3D shape retrieval visualizations on Pascal3D (L3). Ground Truth (GT) is shown for reference.}
\label{fig:additional_retrieval_vis}
\vspace{-0.2cm}
\end{figure*}

Figure~\ref{fig:additional_retrieval_vis} shows some additional 3D shape retrieval results on Pascal3D (L3). PASR not only retrieves the best-matching 3D shapes, but also estimates the pose of each shape accurately.

%% file: main.bib
@String(CVPR= {IEEE Conf. Comput. Vis. Pattern Recog.})

@String(ICCV= {Int. Conf. Comput. Vis.})

@String(ICLR = {Int. Conf. Learn. Represent.})

@String(AAAI = {AAAI})

@String(CVPR  = {CVPR})

@String(ICCV  = {ICCV})

@String(ICLR  = {ICLR})

@article{tangelder2004survey,
  title={A survey of content based 3D shape retrieval methods},
  author={Tangelder, Johan WH and Veltkamp, Remco C},
  journal={Proceedings Shape Modeling Applications, 2004.},
  pages={145--156},
  year={2004},
  publisher={IEEE}
}

@article{xiao2020survey,
  title={A survey on deep geometry learning: From a representation perspective},
  author={Xiao, Yun-Peng and Lai, Yu-Kun and Zhang, Fang-Lue and Li, Chunpeng and Gao, Lin},
  journal={Computational Visual Media},
  volume={6},
  number={2},
  pages={113--133},
  year={2020},
  publisher={TUP}
}

@article{yuille2006vision,
  title={Vision as Bayesian inference: analysis by synthesis?},
  author={Yuille, Alan and Kersten, Daniel},
  journal={Trends in cognitive sciences},
  volume={10},
  number={7},
  pages={301--308},
  year={2006},
  publisher={Elsevier}
}

@misc{liu2023openshape,
      title={OpenShape: Scaling Up 3D Shape Representation Towards Open-World Understanding}, 
      author={Minghua Liu and Ruoxi Shi and Kaiming Kuang and Yinhao Zhu and Xuanlin Li and Shizhong Han and Hong Cai and Fatih Porikli and Hao Su},
      year={2023},
      eprint={2305.10764},
      archivePrefix={arXiv},
      primaryClass={cs.CV}
}

@inproceedings{zhou2023uni3d,
  title={Uni3d: Exploring unified 3d representation at scale},
  author={Zhou, Junsheng and Wang, Jinsheng and Ma, Baorui and Liu, Yu-Shen and Huang, Tiejun and Wang, Xinlong},
  booktitle={International Conference on Learning Representations (ICLR)},
  year={2024}
}

@inproceedings{xue2023ulip,
  title={Ulip: Learning a unified representation of language, images, and point clouds for 3d understanding},
  author={Xue, Le and Gao, Mingfei and Xing, Chen and Mart{\'\i}n-Mart{\'\i}n, Roberto and Wu, Jiajun and Xiong, Caiming and Xu, Ran and Niebles, Juan Carlos and Savarese, Silvio},
  booktitle={Proceedings of the IEEE/CVF conference on computer vision and pattern recognition},
  pages={1179--1189},
  year={2023}
}

@inproceedings{xue2024ulip,
  title={Ulip-2: Towards scalable multimodal pre-training for 3d understanding},
  author={Xue, Le and Yu, Ning and Zhang, Shu and Panagopoulou, Artemis and Li, Junnan and Mart{\'\i}n-Mart{\'\i}n, Roberto and Wu, Jiajun and Xiong, Caiming and Xu, Ran and Niebles, Juan Carlos and others},
  booktitle={Proceedings of the IEEE/CVF Conference on Computer Vision and Pattern Recognition},
  pages={27091--27101},
  year={2024}}

@article{fu2020hard,
  title={Hard example generation by texture synthesis for cross-domain shape similarity learning},
  author={Fu, Huan and Li, Shunming and Jia, Rongfei and Gong, Mingming and Zhao, Binqiang and Tao, Dacheng},
  journal={Advances in neural information processing systems},
  volume={33},
  pages={14675--14687},
  year={2020}
}

@inproceedings{lin2021single,
  title={Single Image 3D Shape Retrieval via Cross-Modal Instance and Category Contrastive Learning},
  author={Lin, Ming-Xian and Yang, Jie and Wang, He and Lai, Yu-Kun and Jia, Rongfei and Zhao, Binqiang and Gao, Lin},
  booktitle={Proceedings of the International Conference on Computer Vision (ICCV)},
  pages={11405--11415},
  year={2021}
}

@article{wu2023generalizing,
    author  = {Wu, Qirui and Ritchie, Daniel and Savva, Manolis and Chang, Angel.X},
    title   = {{Generalizing Single-View 3D Shape Retrieval to Occlusions and Unseen Objects}},
    year    = {2023},
    eprint  = {2401.00405},
    archivePrefix   = {arXiv},
    primaryClass    = {cs.CV}
}

@inproceedings{grabner2019location,
  title={Location field descriptors: Single image 3d model retrieval in the wild},
  author={Grabner, Alexander and Roth, Peter M and Lepetit, Vincent},
  booktitle={2019 international conference on 3d vision (3DV)},
  pages={583--593},
  year={2019},
  organization={IEEE}
}

@article{songsc-ibsr,
author = {Song, Dan and Huo, Shumeng and Fu, Xinwei and Zhang, Chumeng and Li, Wenhui and Liu, An-An},
title = {Cross-Modal Contrastive Learning with a Style-Mixed Bridge for Single Image 3D Shape Retrieval},
year = {2024},
journal = {ACM Trans. Multimedia Comput. Commun. Appl.},
}

@inProceedings{lin2021cmic,
	title={Single Image 3D Shape Retrieval via Cross-Modal Instance and Category Contrastive Learning},
	author={Lin, Ming-Xian and Yang, Jie and Wang, He and Lai, Yu-Kun and Jia, Rongfei and Zhao, Binqiang and Gao, Lin},
	year={2021},
	booktitle={International Conference on Computer Vision (ICCV)}
}

@inproceedings{ma2022robust,
  title={Robust category-level 6d pose estimation with coarse-to-fine rendering of neural features},
  author={Ma, Wufei and Wang, Angtian and Yuille, Alan and Kortylewski, Adam},
  booktitle={European Conference on Computer Vision},
  pages={492--508},
  year={2022},
  organization={Springer}
}

@inproceedings{kortylewski2020compositional,
  title={Compositional convolutional neural networks: A deep architecture with innate robustness to partial occlusion},
  author={Kortylewski, Adam and He, Ju and Liu, Qing and Yuille, Alan L},
  booktitle={Proceedings of the IEEE/CVF Conference on Computer Vision and Pattern Recognition},
  pages={8940--8949},
  year={2020}
}

@article{wang2021nemo,
  title={Nemo: Neural mesh models of contrastive features for robust 3d pose estimation},
  author={Wang, Angtian and Kortylewski, Adam and Yuille, Alan},
  journal={arXiv preprint arXiv:2101.12378},
  year={2021}
}

@article{chen2003visual,
  title={On visual similarity based 3D model retrieval},
  author={Chen, Ding-Yun and Tian, Xiao-Pei and Shen, Yu-Te and Ouhyoung, Ming},
  journal={Computer Graphics Forum},
  volume={22},
  number={3},
  pages={223--232},
  year={2003},
  publisher={Blackwell Publishing, Inc}
}

@article{han2019aggregating,
  title={3D2SeqViews: Aggregating Sequential Views for 3D Global Feature Learning by CNN with Hierarchical Attention Aggregation},
  author={Han, Zhizhong and Zhang, Xiyang and Chen, Chuhang and Han, Junwei and Schwing, Alexander G and Liu, Jingtuo},
  journal={IEEE Transactions on Image Processing},
  volume={28},
  number={8},
  pages={3986--3999},
  year={2019},
  publisher={IEEE}
}

@inproceedings{jiang2019mlvcnn,
  title={MLVCNN: Multi-loop-view Convolutional Neural Network for 3D Shape Retrieval},
  author={Jiang, Jianwen and Xie, Yiqun and Li, Lei and Liu, Xin},
  booktitle={Proceedings of the AAAI Conference on Artificial Intelligence},
  volume={33},
  number={1},
  pages={8513--8520},
  year={2019}
}

@inproceedings{nguyen2022templates,
  title={Templates for 3D Object Pose Estimation Revisited: Generalization to New Objects and Robustness to Occlusions},
  author={Nguyen, Van Nguyen and Hu, Yinlin and Xiao, Yang and Salzmann, Mathieu and Lepetit, Vincent},
  booktitle={Proceedings of the Conference on Computer Vision and Pattern Recognition (CVPR)},
  pages={6771--6780},
  year={2022}
}

@article{ma2024imagenet3d,
  title={ImageNet3D: Towards General-Purpose Object-Level 3D Understanding},
  author={Ma, Wufei and Zhang, Guofeng and Liu, Qihao and Zeng, Guanning and Kortylewski, Adam and Liu, Yaoyao and Yuille, Alan},
  journal={Advances in Neural Information Processing Systems},
  volume={38},
  year={2024}
}

@inproceedings{he2016deep,
  title={Deep residual learning for image recognition},
  author={He, Kaiming and Zhang, Xiangyu and Ren, Shaoqing and Sun, Jian},
  booktitle={Proceedings of the IEEE conference on computer vision and pattern recognition},
  pages={770--778},
  year={2016}
}

@inproceedings{liu2021swin,
  title={Swin transformer: Hierarchical vision transformer using shifted windows},
  author={Liu, Ze and Lin, Yutong and Cao, Yue and Hu, Han and Wei, Yixuan and Zhang, Zheng and Lin, Stephen and Guo, Baining},
  booktitle={Proceedings of the IEEE/CVF international conference on computer vision},
  pages={10012--10022},
  year={2021}
}

@inproceedings{sun2018pix3d,
  title={Pix3d: Dataset and methods for single-image 3d shape modeling},
  author={Sun, Xingyuan and Wu, Jiajun and Zhang, Xiuming and Zhang, Zhoutong and Zhang, Chengkai and Xue, Tianfan and Tenenbaum, Joshua B and Freeman, William T},
  booktitle={Proceedings of the IEEE conference on computer vision and pattern recognition},
  pages={2974--2983},
  year={2018}
}

@inproceedings{xiang2014beyond,
  title={Beyond pascal: A benchmark for 3d object detection in the wild},
  author={Xiang, Yu and Mottaghi, Roozbeh and Savarese, Silvio},
  booktitle={IEEE winter conference on applications of computer vision},
  pages={75--82},
  year={2014},
  organization={IEEE}
}

@article{simeoni2025dinov3,
  title={Dinov3},
  author={Sim{\'e}oni, Oriane and Vo, Huy V and Seitzer, Maximilian and Baldassarre, Federico and Oquab, Maxime and Jose, Cijo and Khalidov, Vasil and Szafraniec, Marc and Yi, Seungeun and Ramamonjisoa, Micha{\"e}l and others},
  journal={arXiv preprint arXiv:2508.10104},
  year={2025}
}

@article{qian2022pointnext,
  title={Pointnext: Revisiting pointnet++ with improved training and scaling strategies},
  author={Qian, Guocheng and Li, Yuchen and Peng, Houwen and Mai, Jinjie and Hammoud, Hasan and Elhoseiny, Mohamed and Ghanem, Bernard},
  journal={Advances in neural information processing systems},
  volume={35},
  pages={23192--23204},
  year={2022}
}

@inproceedings{Peng2023OpenScene,
  title     = {OpenScene: 3D Scene Understanding with Open Vocabularies},
  author    = {Peng, Songyou and Genova, Kyle and Jiang, Chiyu "Max" and Tagliasacchi, Andrea and Pollefeys, Marc and Funkhouser, Thomas},
  booktitle = {Proceedings of the IEEE/CVF Conference on Computer Vision and Pattern Recognition (CVPR)},
  year      = {2023}
}

@inproceedings{wu2025sonata,
    title={Sonata: Self-Supervised Learning of Reliable Point Representations},
    author={Wu, Xiaoyang and DeTone, Daniel and Frost, Duncan and Shen, Tianwei and Xie, Chris and Yang, Nan and Engel, Jakob and Newcombe, Richard and Zhao, Hengshuang and Straub, Julian},
    booktitle={CVPR},
    year={2025}
}

@inproceedings{wu2024ppt,
    title={Towards Large-scale 3D Representation Learning with Multi-dataset Point Prompt Training},
    author={Wu, Xiaoyang and Tian, Zhuotao and Wen, Xin and Peng, Bohao and Liu, Xihui and Yu, Kaicheng and Zhao, Hengshuang},
    booktitle={CVPR},
    year={2024}
}

@article{hegde2023clip,
 title={CLIP goes 3D: Leveraging Prompt Tuning for Language Grounded 3D Recognition},
 author={Hegde, Deepti and Valanarasu, Jeya Maria Jose and Patel, Vishal M},
 journal={arXiv preprint arXiv:2303.11313},
 year={2023}
}

@INPROCEEDINGS{partslip,
  author={Liu, Minghua and Zhu, Yinhao and Cai, Hong and Han, Shizhong and Ling, Zhan and Porikli, Fatih and Su, Hao},
  booktitle={2023 IEEE/CVF Conference on Computer Vision and Pattern Recognition (CVPR)}, 
  title={PartSLIP: Low-Shot Part Segmentation for 3D Point Clouds via Pretrained Image-Language Models}, 
  year={2023},
  volume={},
  number={},
  pages={21736-21746},
  doi={10.1109/CVPR52729.2023.02082}
}

@article{zhang2021pointclip,
  title={PointCLIP: Point Cloud Understanding by CLIP},
  author={Zhang, Renrui and Guo, Ziyu and Zhang, Wei and Li, Kunchang and Miao, Xupeng and Cui, Bin and Qiao, Yu and Gao, Peng and Li, Hongsheng},
  journal={arXiv preprint arXiv:2112.02413},
  year={2021}
}

@article{Zhu2022PointCLIPV2,
    title={PointCLIP V2: Prompting CLIP and GPT for Powerful 3D Open-world Learning},
    author={Zhu, Xiangyang and Zhang, Renrui and He, Bowei and Guo, Ziyu and Zeng, Ziyao and Qin, Zipeng and Zhang, Shanghang and Gao, Peng},
    journal={arXiv preprint arXiv:2211.11682},
    year={2022},
}

@article{zhang2025concerto,
  title={Concerto: Joint 2D-3D Self-Supervised Learning Emerges Spatial Representations},
  author={Zhang, Yujia and Wu, Xiaoyang and Lao, Yixing and Wang, Chengyao and Tian, Zhuotao and Wang, Naiyan and Zhao, Hengshuang},
  journal={arXiv preprint arXiv:2510.23607},
  year={2025}
}

@inproceedings{wimmer2024back,
  title={Back to 3d: Few-shot 3d keypoint detection with back-projected 2d features},
  author={Wimmer, Thomas and Wonka, Peter and Ovsjanikov, Maks},
  booktitle={Proceedings of the IEEE/CVF conference on computer vision and pattern recognition},
  pages={4154--4164},
  year={2024}
}

@InProceedings{Yuan_2025_ICCV,
    author    = {Yuan, Xiaoding and Zhang, Guofeng and Kaushik, Prakhar and Jesslen, Artur and Kortylewski, Adam and Yuille, Alan},
    title     = {Scaling 3D Compositional Models for Robust Classification and Pose Estimation},
    booktitle = {Proceedings of the IEEE/CVF International Conference on Computer Vision (ICCV)},
    month     = {October},
    year      = {2025},
    pages     = {6406-6415}
}

@inproceedings{shi2025chain,
  title={Chain of Semantics Programming in 3D Gaussian Splatting Representation for 3D Vision Grounding},
  author={Shi, Jiaxin and Xiang, Mingyue and Sun, Hao and Huang, Yixuan and Weng, Zhi},
  booktitle={Proceedings of the Computer Vision and Pattern Recognition Conference},
  pages={24560--24569},
  year={2025}
}

@article{ma2025spatialreasoner,
  title={Spatialreasoner: Towards explicit and generalizable 3d spatial reasoning},
  author={Ma, Wufei and Chou, Yu-Cheng and Liu, Qihao and Wang, Xingrui and de Melo, Celso and Xie, Jianwen and Yuille, Alan},
  journal={arXiv preprint arXiv:2504.20024},
  year={2025}
}

@article{chen2025splat,
  title={Splat-nav: Safe real-time robot navigation in gaussian splatting maps},
  author={Chen, Timothy and Shorinwa, Ola and Bruno, Joseph and Swann, Aiden and Yu, Javier and Zeng, Weijia and Nagami, Keiko and Dames, Philip and Schwager, Mac},
  journal={IEEE Transactions on Robotics},
  year={2025},
  publisher={IEEE}
}

@article{yan2024robotron,
  title={RoboTron-Mani: All-in-One Multimodal Large Model for Robotic Manipulation},
  author={Yan, Feng and Liu, Fanfan and Zheng, Liming and Zhong, Yufeng and Huang, Yiyang and Guan, Zechao and Feng, Chengjian and Ma, Lin},
  journal={arXiv preprint arXiv:2412.07215},
  year={2024}
}

@inproceedings{jesslen2024novum,
  title={Novum: Neural object volumes for robust object classification},
  author={Jesslen, Artur and Zhang, Guofeng and Wang, Angtian and Ma, Wufei and Yuille, Alan and Kortylewski, Adam},
  booktitle={European Conference on Computer Vision},
  pages={264--281},
  year={2024},
  organization={Springer}
}

@article{hu2023cross,
  title={Cross-domain image-object retrieval based on weighted optimal transport},
  author={Hu, Nian and Huang, Xiangdong and Li, Wenhui and Li, Xuanya and Liu, An-An},
  journal={IEEE Transactions on Multimedia},
  volume={25},
  pages={9557--9571},
  year={2023},
  publisher={IEEE}
}

@inproceedings{pal2024domain,
  title={Domain Adaptive 3D Shape Retrieval from Monocular Images},
  author={Pal, Harsh and Khandelwal, Ritwik and Pande, Shivam and Banerjee, Biplab and Karanam, Srikrishna},
  booktitle={Proceedings of the IEEE/CVF Winter Conference on Applications of Computer Vision},
  pages={3192--3201},
  year={2024}
}

@article{song2023domain,
  title={Domain-specific modeling and semantic alignment for image-based 3D model retrieval},
  author={Song, Dan and Jiang, Xue-Jing and Zhang, Yue and Zhang, Fang-Lue and Jin, Yao and Zhang, Yun},
  journal={Computers \& Graphics},
  volume={115},
  pages={25--34},
  year={2023},
  publisher={Elsevier}
}

@article{song2025adaptive,
  title={Adaptive CLIP for open-domain 3D model retrieval},
  author={Song, Dan and Qiang, Zekai and Zhang, Chumeng and Wang, Lanjun and Liu, Qiong and Yang, You and Liu, An-An},
  journal={Information Processing \& Management},
  volume={62},
  number={2},
  pages={103989},
  year={2025},
  publisher={Elsevier}
}

@ARTICLE{10818713,
  author={Dai, Yue and Feng, Yifan and Ma, Nan and Zhao, Xibin and Gao, Yue},
  journal={IEEE Transactions on Pattern Analysis and Machine Intelligence}, 
  title={Cross-Modal 3D Shape Retrieval via Heterogeneous Dynamic Graph Representation}, 
  year={2025},
  volume={47},
  number={4},
  pages={2370-2387},
  doi={10.1109/TPAMI.2024.3524440}}

@article{chu2024open,
  title={OPEN: Occlusion-invariant perception network for single image-based 3D shape retrieval},
  author={Chu, Fupeng and Cong, Yang and Chen, Ronghan},
  journal={IEEE Transactions on Circuits and Systems for Video Technology},
  volume={34},
  number={9},
  pages={7998--8012},
  year={2024},
  publisher={IEEE}
}

@misc{oquab2023dinov2,
  title={DINOv2: Learning Robust Visual Features without Supervision},
  author={Oquab, Maxime and Darcet, Timothée and Moutakanni, Theo and Vo, Huy V. and Szafraniec, Marc and Khalidov, Vasil and Fernandez, Pierre and Haziza, Daniel and Massa, Francisco and El-Nouby, Alaaeldin and Howes, Russell and Huang, Po-Yao and Xu, Hu and Sharma, Vasu and Li, Shang-Wen and Galuba, Wojciech and Rabbat, Mike and Assran, Mido and Ballas, Nicolas and Synnaeve, Gabriel and Misra, Ishan and Jegou, Herve and Mairal, Julien and Labatut, Patrick and Joulin, Armand and Bojanowski, Piotr},
  journal={arXiv:2304.07193},
  year={2023}
}

@article{duenkel2025diysc,
    title = {Do It Yourself: Learning Semantic Correspondence from Pseudo-Labels},
    author = {D{\"u}nkel, Olaf and Wimmer, Thomas and Theobalt, Christian and Rupprecht, Christian and Kortylewski, Adam},
    journal = {arXiv preprint arXiv:2506.05312},
    year = {2025}
  }
